%% file: main.tex
\documentclass[twoside,11pt]{article} 
\usepackage{jmlr2e} 

\usepackage{amsmath}                       
\usepackage{makecell}                      
\usepackage{tikz-cd}                       
\usepackage[boxruled]{algorithm2e}         

\makeatletter
\newcommand{\algorithmfootnote}[2][\footnotesize]{%
  \let\old@algocf@finish\@algocf@finish
  \def\@algocf@finish{\old@algocf@finish
    \leavevmode\rlap{\begin{minipage}{\linewidth}
    #1#2
    \end{minipage}}%
  }%
}
\makeatother

\usepackage{todonotes}

\usepackage[makeroom]{cancel} 

\begin{document}
\title{Unsupervised Recalibration\thanks{Research supported by GitHub.}
} 

\ShortHeadings{Unsupervised Recalibration}{Ziegler and Czy\.z}


\editor{---}

\author{\name Albert\ Ziegler \email wunderalbert@github.com \\
        \addr GitHub, Blue Boar Court\\
        9 Alfred Street\\
        Oxford OX1 4EH, United Kingdom
        \AND \name Pawe\l\ Czy\.z \email pawel.czyz@st-hughs.ox.ac.uk \\
        \addr St Hugh's College, University of Oxford,\\
        St Margaret's Road,\\
        Oxford OX2 6LE, United Kingdom
}

\maketitle              

\input{sec-abstract}

\input{sec-introduction}

\input{own-commands}
\input{sec-maths}

\input{sec-experiment}

\input{sec-afterword}

\bibliography{literature}

\appendix
\input{sec-appendix}



\end{document}

%% file: sec-abstract.tex
\begin{abstract}
Unsupervised recalibration (URC) is a general way to improve the accuracy of an already trained probabilistic classification or regression model upon encountering new data while deployed in the field. URC does not require any ground truth associated with the new field data. URC merely observes the model's predictions, and recognizes when the training set is not representative of field data by noting a divergence from the expected distribution of predictions. It works backwards to determine the magnitude of the bias and then removes it.

URC can be particularly useful when applied separately to different subpopulations observed in the field that were not considered as features when training the machine learning model. This makes it possible to exploit subpopulation information without retraining the model or even having ground truth for some or all subpopulations available. 

Additionally, if these subpopulations are the object of study, URC serves to determine the correct ground truth distributions for them, where naive aggregation methods, like averaging the model's predictions, systematically underestimate their differences.



\end{abstract}


%% file: sec-introduction.tex
\section{Introduction}

The life cycle of a classification or regression model normally consists of two distinct phases:

\begin{enumerate}
    \item In the \textbf{development phase}, we select the model's architecture, tweak its parameters and possibly evaluate it according to some \textbf{input data}\footnote{This includes training data as well as any validation, test or holdout data.}. Ground truth for this training data is known. This task is usually performed by a ML engineer in their lab. When they evaluate the (partially or fully constructed) model, they do so in order to learn more about the model.
    \item In the \textbf{application phase}, the model is fixed and gets applied to some \textbf{field data}. Ground truth for that field data is unknown. This task might still be performed by an ML engineer, or alternatively by a field operative with no background in how the model works. When they evaluate the model, they do so in order to learn more about the world.
\end{enumerate}

These phases may be alternated iteratively, sometimes in rapid succession like in online learning, in order to update the model as more input data becomes available. In some applications, field data of the previous iteration will in time get augmented by ground truth to become the input data of the subsequent iteration. In other applications, the application phase may be completely decoupled from the construction phase, for example if model deployment consists in the publication of a classification technique in a textbook.

\subsection{Learning during application}

Most of machine learning literature focuses on the development phase, helping ML engineers to fashion or update models that fit the ground truth in a generalizable way. But in practice, many models spend most of their lifetime in the application phase, and just observing the model's predictions in this phase is informative and can lead us to improve the model without any need to access ground truth.

In particular, very often we are able to observe the model's predictions in the application phase differing slightly according to some categorical property of the samples that was not used as a feature for the model. This may have been due to any of the following reasons, and often a combination of them:
\begin{itemize}
    \item Some of the categories in the field data were not represented in the input data.
    \item The relevance of the property in question was not anticipated during development.
    \item There was insufficient input data for (some of) the different categories, and some may have been missing from the input data altogether.
    \item The property was unavailable at development time, or measuring would have been too resource intensive.
    \item The model was developed to function in a more general setting where the categories may not be available.
\end{itemize}

A typical example would be a fast-but-imprecise medical test. During application, practitioners might identify possible risk factors for which the model's predictions are, on average, higher. A gold standard test could help quantify the influence of those risk factors, but is often too resource expensive to use on a large scale.

There are two possible reasons for this observed difference in predictions: either the different categories have different ground truth distributions, or the relationship between ground truth and model predictions is fundamentally different for the different categories. We contend that in many cases, it is reasonable to assume that the former is the driving factor behind the observed differences between categories (Axiom \ref{ax:consistency}).


However, a straightforward extension to the model to pull in the property in question is not usually possible, for any of the following reasons, and often a combination of them:
\begin{itemize}
    \item Ground truth for field data may not be known.
    \item The classifier may be a black box (e.g. because it was produced by a third party).
    \item The classifier is of such a form that it may not easily be extended with new features.
    \item Re-training the classifier is too difficult, either conceptionally or computationally.
    \item The amount of data is insufficient for some or all categories.
\end{itemize}


Nevertheless, we will often observe classifier predictions for samples from the same category to be more homogeneous than all predictions as a whole. This indicates a correlation between ground truth and category. Because the model has untapped potential regarding this categorical property, it will tend to underestimate the actual differences (Theorem \ref{thm:naiveunderestimates}). By estimating and accounting for the real correlation, classifier predictions on individual samples as well as on the group as a whole can be improved. 

Since the correlation is underestimated without specific post-processing, a particularly important application of URC is when the differences between the categories are the main focus of the model's current application.



In the medical example above, the model may be used to quantify the possible risk factors. Using the naive estimation of risk factors by taking the average model prediction for each category as an estimator for the incidence rate would underestimate the influence of the category. 





\subsection{Relationship to existing techniques}
Calibrating a classifier in order to make its predictions match the actual distribution of ground truth is an established post-processing technique \citep{calibration}. It consists of tweaking the classifier predictions in a generalizable way to match the observed ground truth distribution in each ``case'', where where the cases usually are a group of samples with similar predictions. Our technique can be seen as a kind of calibration, albeit one that occurs during the application phase, where the ground truth distribution can not be directly observed.

We propose to see the emergence of an unaccounted category during the application phase as an example of \emph{data shift} \citep{Moreno-Torres}, which occurs independently for each level in the new category.
To account for this data shift, a \emph{quantification} problem (see e.g. \citet{gonzalez-review-quantification} for an overview) needs to be solved for each category. This refers to the task of estimating the ground truth distribution in an unlabeled set (a category during the application phase) from the distribution in a labeled data set (from the development phase). Unsupervised recalibration can therefore be considered as another quantification technique -- we compare it with established methods in \textsection\ref{sec:experiments}.

\section{Outline}
\label{section:outline}
In \textsection\ref{sec:global_unsupervised_calibration}, we introduce the heart of URC: a technique to recalibrate a classifier that was calibrated on a biased input set without direct information about that bias (``global unsupervised recalibration''). In \textsection\ref{sec:local_unsupervised_calibration} we make use of this technique to improve general classifiers based on context (``local unsupervised recalibration''). Together, these sections form Algorithm \ref{algo:urc_class}.

In \textsection\ref{sec:regression}, we will discuss how to extend this approach to regression problems. In \textsection\ref{sec:experiments} we show how URC can improve a  classifier's performance in practice and we compare URC with three existing quantification algorithms. We close with a discussion of when \textit{not} to use URC in \textsection\ref{sec:counterindications}.

%% file: own-commands.tex
%
%
%
%
%
%
%
%
%
%
%
%

\newcommand{\func}[3]{#1\colon #2\to #3} 
\newcommand{\dev}{\text{dev}} 
\newcommand{\app}{\text{app}} 
\newcommand{\noth}{} 

\newcommand{\MockProb}[2]{P_{#1}\left(#2\right)} 
\newcommand{\MockEV}[2]{\mathbb E_{#1}\left(#2\right)}
\makeatletter
\newcommand{\Prob}[2][\@empty]{%
    \MockProb{\ifx\@empty#1\relax\noth\else#1\fi}{#2}
}
\newcommand{\CondProb}[3][\@empty]{%
    \Prob[#1]{#2\,|\,#3}
}

\newcommand{\EV}[2][\@empty]{%
    \MockEV{\ifx\@empty#1\relax\noth\else#1\fi}{#2}
}

\makeatother

\newcommand{\Reals}{\mathbb R}
\newcommand{\assuming}{\,|\,}

\newcommand{\range}[1]{1, 2, \dots, #1}
\newcommand{\pred}[1]{\mathrm{pred}\,#1}
\newcommand{\refeq}[1]{(\ref{#1})}

%% file: sec-maths.tex
\section{Global unsupervised recalibration}\label{sec:global_unsupervised_calibration}

	Consider a space of \textbf{features} $\mathcal X$ and space of \textbf{labels} $\mathcal Y=\{\range n\}$. We are interested in a \textbf{classifier} $\func f {\mathcal X}{\Reals^n}$ predicting probabilities of labels $\mathcal Y$, i.e. if $x\in \mathcal X$, then we want the $k$-th component of the $f(x)$ vector to represent the chance that $y=k$.
	
	Both for classifier development and in the application phase, samples $(x, y)\in\mathcal{X}\times\mathcal{Y}$ are drawn i.i.d. from a probability distribution on $\Omega = \mathcal{X}\times\mathcal{Y}$. 
	
	We will denote the development distribution by $P_\dev$ and the application distribution by $P_\app$. We assume $P_\app$ to be absolutely continuous with respect to $P_\dev$.
	
    On the space $\Omega = \mathcal{X}\times\mathcal{Y}$ we define the following random variables:
\begin{itemize}
	\item $\func X \Omega {\mathcal X}$ is the projection giving features,
	\item $\func Y \Omega {\mathcal Y}$ is the projection giving labels,
	\item $\func C \Omega {\Reals^n}$ is the random variable representing $f$, i.e. $C = f\circ X$
		\begin{center}
			\begin{tikzcd}
				\Omega \arrow[rd, "C"] \arrow[d, "X", swap] \\
				\mathcal X \arrow[r, "f", swap]& \mathbb R^n
			\end{tikzcd}
		\end{center}
		By slight abuse of terminology we will refer to this random variable giving the classifier predictions as the \textbf{classifier} itself,
	\item $\func{C_k}{\Omega}{\Reals}$ is the $k$-th component of $C$. 
\end{itemize}

\begin{remark}
    We assume that the projections $X, Y$ and the classifier $C$ are measurable. This is enough to guarantee that all functions and sets considered below are measurable. 
    
    Without loss of generality we also assume $\mathcal{Y}$ to only observe labels that are observable in principle, i.e. $\Prob[\dev]{Y=i}>0$ for all $i\in\mathcal{Y}$.
\end{remark}

We say that a classifier is \textbf{calibrated on the training set} if $C_i$ describes the probability of $Y=i$ conditioned on $C$ , i.e. the following two functions from $\Omega$ to $\Reals$ are identical:
\begin{equation}\label{eq:cal_for_p}
    C_i = \CondProb[\dev]{Y=i}{C}
\end{equation}
As usual, the conditional probability of an event $Y=i$ conditioned on a random variable is the conditional expectation of the indicator function $1_{Y=i}$ conditioned on the random variable (see for example \citet{conditional-expectation}).

We call the training population \textbf{unbiased} for $Y$ if

\begin{equation}\label{eq:unbiased}
	\Prob[\dev]{Y=i} = \Prob[\app]{Y=i}.
\end{equation}

We would like to have a classifier that is \textbf{calibrated overall}, i.e.
\begin{equation}\label{eq:cal_overall}
	C_i = \CondProb[\app]{Y=i}{C}.
\end{equation} 

\subsection{From training to practice}\label{ss:generalization_issues}
There are established methods to calibrate a classifier on the training set (eq. \refeq{eq:cal_for_p} holds, at least in approximation), and we will assume the classifier to be calibrated on the training set in the following. But such a classifier may still not be calibrated overall (eq. \refeq{eq:cal_overall} does not hold) for two reasons.

    First, classifier $C$ may pick up features specific to the training population only. There are various techniques reducing overfitting \citep{invariant-risk-minimization, overfitting} and we assume that the classifier $C$ was trained in a way that this is not a major problem.

    Second, the training population may be biased in some way, i.e. equation \refeq{eq:unbiased} does not hold. This is common, since the training population often comes from a different source (where labels were available). Often, this is even deliberate in order to facilitate training, for example through stratification \citep{stratification}. 

\begin{example}[Entomologist's classifier]\label{ex:butterflies_setup}
    Consider a classifier used to classify butterflies and beetles. As a simple feature space one can use wingspan and body weight. For simplicity assume that each of these features can take two values: ``small'' or ``large''.

    Equation \refeq{eq:cal_for_p} states that a classifier was calibrated on the training data set. This means that if all we know is that a sample is from the training set and the classifier says ``beetle with 90\%'' (left hand side of eq. \refeq{eq:cal_for_p}), this chance is actually 90\% (right hand side). E.g. the classifier might only say this for ``small wings, large body", then 90\% of ``small wings, large body'' samples from the training set are actually beetles.

    If the training samples had been collected in a forest, beetles would be overrepresented compared to, say, a meadow. So the classifier is biased (eq. \refeq{eq:unbiased} is does not hold): the chance to have a beetle in the training set (left hand side) is higher than overall (right hand side).

    Thus also equation \refeq{eq:cal_overall} does not hold: $C_\text{beetle}$ is be too large and $C_\text{butterfly}$ is too low.
\end{example}

Such a bias is not just a standard miscalibration problem, but a (bias) quantification problem: the classifier is not merely overconfident \citep{platt-scaling} or underconfident \citep{beta-calibration}, but systematically underestimates the probability of $Y=i$ in case %
$\Prob[\app]{Y=i} > \Prob[\dev]{Y=i}$  
or systematically overestimates it in case %
$\Prob[\app]{Y=i} < \Prob[\dev]{Y=i}$. 

Biased training sets are a known problem, and previous approaches aimed at quantifying and minimising that problem. For example, \citep{biasthroughstratification} suggest a sample stratification strategy that yields the minimax error considering an unknown bias. However, this error is still considerable, which is why we will aim at correcting it entirely.

\subsection{Consistency assumption}\label{ss:consistency}
We want to account for the training data being biased ($P_\dev \neq P_\app$), but if $P_\dev$ and $P_\app$ are completely unrelated, there is no point in learning. We make the following \textbf{Consistency assumption} that the bias is ``tame'':
\begin{center}
\begin{minipage}{0.75\textwidth}
    \indent Although the sample may be biased with respect to $Y$, the sample does not differ from the overall population with respect how $Y$ and $C$ interact --- conditioned on each class $Y = i$, the distribution of $C$ is assumed to be identical in the training set and the real world.
\end{minipage}
\end{center}
In other words, we assume that Axiom \ref{ax:consistency} holds.
\begin{axiom}\label{ax:consistency}
    Let $\mathcal A$ be the $\sigma$-algebra generated by $C$. For every event $\alpha\in \mathcal A$:
    \begin{equation}\label{eq:P_unbiased_for_C_rel_X}
        \CondProb[\dev]{\alpha}{Y=i} = \CondProb[\app]{\alpha}{Y=i}
    \end{equation}
\end{axiom}

 In many applications, this assumption is highly plausible since it already holds for the features $X$ used to compute $C$ -- for each class $Y=i$, the distribution of the features for members of this class does not differ between training set and overall population. In other words, being class $i$ is associated with the same features in both the development and application, it's just that the class itself is more common in one and rarer in another.
 
 In particular, this is to be expected if the training set is obtained through stratified sampling. The consistency assumption is ubiquitous in quantification literature under different names, as ``prior probability shift'' or ``global shift'', cf. \citet[p. 3]{Vaz-Izbicki-Stern}, \citet[p. 24f.]{Saerens-2001-adjustingtheoutputs}, \citet[p. 6]{tasche-fisher-consistency}.

\begin{example}[Entomologist's classifier, continued]\label{ex:butterflies_consistency}
    
    We have to assume that the abundance of butterflies differs significantly between development in the forest and application in the meadow. It seems far less risky to assume that butterflies from the forest look similar to butterflies from the meadow, they are just rarer. 
    
    Since the distribution of features is assumed to be the same for butterflies from meadow and forest, and the classifier is a function of the features, the distribution of classifier predictions for butterflies during classifier development (left hand side of eq. \refeq{eq:P_unbiased_for_C_rel_X}) must also be the same as the distribution of classifier predictions for butterflies during classifier application (right hand side).
\end{example}

\subsection{Global unsupervised recalibration for known bias}\label{ss:knownbias}
If the bias is known, recalibration becomes applying a simple formula.

\begin{lemma}\label{lemma:knownbias}
    Let the following random variables representing unnormalized probabilities be defined as
    \begin{equation}
        \bar p_i := C_i \cdot \frac{\Prob[\app]{Y=i}}{\Prob[\dev]{Y=i}}
    \end{equation}
    Then
    \begin{equation}
        \CondProb[\app]{Y=i}{C} = \frac{\bar{p}_i}{\sum\limits_j{\bar{p}_j}}
    \end{equation}
\end{lemma}
\begin{proof}
    Direct calculation (cf. \textsection 2.2 of \citet{Saerens-2001-adjustingtheoutputs}).
\end{proof}
\subsection{Global unsupervised recalibration for unknown bias}\label{ss:global_recalibration_unknown_bias}
Due to Lemma \ref{lemma:knownbias}, we need a sensible estimate for the true distribution of $Y$ in the population. A first pass might use the naive estimator:
\begin{definition}
    An unbiased estimator for the expected value $\EV[\app]{C_i}$ is called a \textbf{naive estimator} for $\Prob[app]{Y = i}$.
\end{definition}
Such an estimator is easy to compute, as it only requires averaging the classifier predictions over the field data. 
However, while useful for predicting the \textit{direction} of the bias given enough data, this will normally \textit{underestimate} the magnitude of the bias and will not approach the true value as the sample size increases:

\begin{theorem}\label{thm:naiveunderestimates}
    Consider a binary\footnote{We expect most non-binary classifiers to exhibit similar behaviour, but while all binary classifiers fulfil the theorem, it is possible to construct a counter-example for \emph{ternary} classifiers.} classifier, i.e. $\mathcal Y = \{0, 1\}$ and $C_0+C_1=1$.
    If
    \begin{equation}
        \label{eq:sandwich_assumption}
        0 < \Prob[\app]{Y=1} <  \Prob[\dev]{Y=1},
    \end{equation}
    then
    \begin{equation}\label{eq:sandwich}
        \Prob[\app]{Y=1} < \EV[\app]{C_1} \leq \Prob[\dev]{Y=1}.
    \end{equation}
    The corresponding statement with $Y=0$ also holds.
\end{theorem}

\begin{proof}
    Let $\pi_\app := P_\app \circ C_1^{-1}$ be the pushforward measure. 
    Using Lemma \ref{lemma:knownbias} and equation \refeq{eq:sandwich_assumption}:
    \begin{align*}
        \Prob[\app]{Y = 1}
        &= \int \CondProb[\app]{Y=1}{C_1}\; \mathrm{d}\pi_\app\\
        & = \int \frac{C_1}{
        C_1 + C_0 \cdot \frac{\Prob[\app]{Y=0} \Prob[\dev]{Y=1}}{\Prob[\dev]{Y=0} \Prob[\app]{Y=1} }
        }  \;\mathrm{d}\pi_\app\\
        & < \int \frac{C_1}{
        C_1 + C_0 \cdot 1
        }  \;\mathrm{d}\pi_\app\\
        & = \int C_1\;\mathrm{d}\pi_\app
        = \EV[\app]{C_1}
    \end{align*}
    Where the inequality is strict because $\Prob[\app]{C_0 > 0} > 0$. This holds since otherwise, $\Prob[\app]{Y=1}$ would have to be $1$. This proves the first part of equation \refeq{eq:sandwich}. 
    
    
    For the second part of the inequality, we argue by case distinction over $Y$. To make this exact, we need to define the conditional measures
    \begin{center}
        $\pi_0 := \CondProb[\dev]{\bullet}{Y=0}$ and $\pi_1 := \CondProb[\dev]{\bullet}{Y=1}$.
    \end{center} Note that because of Axiom \ref{ax:consistency}, we might also have used $P_\app$ here.
    
    \begin{align*}
        \Prob[\dev]{Y=1} - \EV[\app]{C_1} &= \int C_1\; \mathrm{d}\pi_\dev -  \int C_1\; \mathrm{d}\pi_\app \\
        &= \hspace{0.7cm} \Bigg( \Prob[\dev]{Y=0}\int C_1\; \mathrm{d}\pi_0 + \Prob[\dev]{Y=1}\int C_1\;\mathrm{d}\pi_1 \Bigg)\\ 
        &\hspace{0.7cm}-\Bigg( \Prob[\app]{Y=0}\int C_1\; \mathrm{d}\pi_0 + \Prob[\app]{Y=1}\int C_1\; \mathrm{d}\pi_1 \Bigg) \\
        &= \bigg(\Prob[\dev]{Y=1}-\Prob[\app]{Y=1}\bigg) \cdot \left( \int C_1 \;\mathrm{d}\pi_1 - \int C_1 \;\mathrm{d}\pi_0 \right)\\
        &= \hspace{0.5cm}\bigg(\Prob[\dev]{Y=1}-\Prob[\app]{Y=1}\bigg) \\
        & \hspace{0.7cm} \cdot \bigg(\EV[\dev]{C_1\assuming Y=1} - \EV[\dev]{C_1 \assuming Y=0}\bigg)
    \end{align*}
    The first term is positive from the assumption and the second is non-negative for every classifier calibrated on the training data set. We also see that equality is strict if the classifier has any discriminative power at all.
    
    
    
    
    
    
\end{proof}

In general, if the classifier is very accurate, i.e. for every $i\in \mathcal Y$ we have $C_i\cdot (1-C_i) \approx 0$, then the naive estimator will be reasonably close to the desired $P(Y = i)$ (and the classifier will already be reasonably well calibrated to the field data). Otherwise, an alternative is needed.

Our alternative centers around partitioning classifier predictions into a finite number of clusters and analysing how often each cluster appears. We will then work backwards to determine which ground truth distribution might have caused these observations. The partition is a hyperparameter of URC.

\begin{definition}
    Call a family of sets $A=(A_i)_{i=\range n}$, where $A_i\subseteq \mathbb R^n$, a \textbf{partition for $C$} if:
    \begin{itemize}
        \item $\Prob{C \in A_i \cap A_j} = 0$ for $i\neq j$ and
        \item $\Prob{C \in \bigcup\limits_{i=1}^n A_i} = 1$.
    \end{itemize}
    for both $P_\dev$ and\footnote{This equation holding for $P_\dev$ implies it also holding for $P_\app$ because of absolute continuity.} $P_\app$.

    For a given partition $A$ of $C$, define  
    \begin{equation}
        M_A = (\CondProb[\dev]{C\in A_j}{Y=i})_{i, j = \range n},
    \end{equation}
    where each element $m_{i,j}$ of $M_A$ is the probability the classifier predicts an element of partition $j$ given an example of category $i$ in the training set. $M_{A}$ therefore encodes the distribution of predictions conditional on the ground truth. 
    
    Define further
    \begin{equation}
        \vec v_A =  (\Prob[\app]{C\in A_j})_{j = \range n},
    \end{equation}
    where $k$-th entry of $\vec v_A$ is the probability the classifier predicts an element of partition $A_k$ in the field.
\end{definition}

The matrix $M_A$ is computed during the development phase. The vector $\vec v_A$ is observed in the field. These two include all the information URC requires to estimate the ground truth distribution in the field data (although a few extra summaries on the training data might be useful for regularisation, see remark \ref{rem:global_urc}).

The URC equations work for any partition\footnote{In fact, they generalize to a partition into more than $n$ sets.} of $C$. For reasons of numerical stability, it is sensible to choose a system of sets with similar probability and low variation. In the binary classification case, we suggest taking intervals which appear as equally likely from the training set. For a partition into $m$ intervals, this would mean for all $i\leq m$:
\begin{equation}\label{eq:example_partition}
    A_i = \left\{(y_1, y_2) \assuming y_1 + y_2 = 1 \wedge \frac{i-1}{m} < \Prob[\dev]{C_1 \leq y_1} \leq \frac{i}{m}\right\}
\end{equation}

\begin{lemma}\label{lemma:solvematrixeq}
    Let $(A_i)_{i=1, 2,..., n}\subseteq\mathbb{R}^n$ be a partition for $C$ and let
        \begin{equation*}
            \vec p_y = (\Prob[\app]{Y=1}, \dots, \Prob[\app]{Y=n}).
        \end{equation*}
    Then
    \begin{equation}
    \label{eq:matrix_multplication}
        M_A \cdot \vec p_y = \vec v_A
    \end{equation}
\end{lemma}

\begin{proof}
    Because of the consistency assumption (equation \refeq{eq:P_unbiased_for_C_rel_X}),
    \begin{equation*}
        \CondProb[\dev]{C\in A_j}{Y=i} = \CondProb[\app]{C\in A_j}{Y=i}.
    \end{equation*}
    So the equation follows from case distinction on $Y=1 \vee ... \vee Y = n$.
\end{proof}

Lemma \ref{lemma:solvematrixeq} is system of linear equations relating the ground truth (which is not directly observable) to the model predictions (which are directly observable). $M_A$ is usually full rank (depending on the choice of partition), and in theory then we could solve for $\vec p_y$ \citep{lipton2018detecting}.

However, we found this to be of limited use in practice:
\begin{itemize}
    \item if the classifier is not very accurate, the condition number of equation \refeq{eq:matrix_multplication} will be very high,
    \item sample limitations may usually leave at least some uncertainty when estimating the probability of $\Prob[\app]{C\in A_j}$ or matrix $M_A$. 
\end{itemize}
Therefore Lemma \ref{lemma:solvematrixeq} often will not provide a precise and accurate estimate of $\vec p_y$ vector in a meaningful way.

Moreover, in the case we extend our approach to partitions with more than $n$ members, this uncertainty will even lead to the (now over-determined) system normally being unsolvable when using the approximate values for $\vec v_A$.

However, it does give rise to an optimization problem. Instead of solving for the vector $\vec p_y$ directly, we can judge how likely a series of observations was for a candidate solution $\vec p_y$. Define a loss function as follows.

\begin{definition}\label{def:Lll}
    For an unbiased sample $S$ of size $|S|$ and a partition $A$ for $C$, define the negative log-likelihood loss as $\func{L_\mathrm{nll}}{[0, 1]^n}{\Reals^+\cup\{\infty\}}$ by
    \begin{equation}
        L_\mathrm{nll}(\vec p) = -\log B(|S|, \pred S, M_A \cdot \vec p),
    \end{equation}
    where $B(m, \vec k, \vec p)$ is the multinomial mass function, i.e.
    \begin{equation}
        B(m, \vec{k}, \vec{p}) =  \binom{n}{k_1, ..., k_n}p_1^{k_1}\cdot...\cdot p_n^{k_n}
    \end{equation}
    and $\pred S = |S\wedge C\in A_i|_{i=\range n}$ is the histogram of classifier predictions.
\end{definition}

\begin{proposition}
    If $M_A$ is full rank, $L_\mathrm{nll}$ has a single global minimum, which for $|S|\rightarrow\infty$ converges against $\vec p_y$.
\end{proposition}
\begin{proof}
    The minimum of the multinomial function  $B(m, \vec k, \vec p)$ is attained at $\vec{p} = \vec{k} / m$, so in the limit case, where 
    \begin{equation}
        \frac{\pred S}{|S|} \rightarrow \Prob[\app]{C\in A_i},
    \end{equation}
    we approach $\vec p_y$, the solution of the equation in Lemma \ref{lemma:solvematrixeq}.
\end{proof}
Solving for minimal $L_\mathrm{nll}$ directly can be risky due to the ill conditioned nature of $M_A$. Moreover $\pred S$ may not be meaningful for small $|S|$.

To overcome this, we add a regularization loss.

\begin{theorem}\label{thm:combinedloss}
    Let $\func{L_\mathrm{reg}}{[0, 1]^n}{\Reals^+ \cup\{\infty\}}$
    be any $C^2$ function such that
    \(
        L_\mathrm{reg}(\vec p_y)<\infty
    \)
    and let $M_A$ be full rank. 

Then the global minimum of $L_\mathrm{nll} + L_\mathrm{reg}$:
\begin{enumerate}
    \item ... exists and is unique with arbitrarily high probability for sufficiently large $|S|$.
    \item ... converges in probability to $\vec p_y$ as $|S|\rightarrow\infty$.
    \item ... is the maximum a posteriori estimate \citep{maximumaposteriori} for $\vec p_y$ for the prior probability proportional to $\exp L_\mathrm{reg}$.
\end{enumerate}
\end{theorem}

\begin{proof}
\begin{enumerate}
    \item Take any $x<1$. We need to find an $k$ such that for $|S|\ge k$, the probability for unique existence of the minimum is at least $x$. Since $L_\mathrm{reg}$ is continuous, there is a compact neighborhood $U_1$ of $\vec p_y$ in which $L_\mathrm{reg}<\infty$. By the extreme value theorem, there is a $b\in\mathbb{R}$ such that \[\frac{\partial ^2 L_\mathrm{reg}}{\partial p_i^2} >b\] for all $i$ and all points in $U_1$. 
    
    The second derivative of the multinomial density $B$ in any direction is bounded from below by $|S|$, a value which is attained for $\vec k  = (|S|, 0, ..., 0)$ and $\vec{p}=(1, 0, ..., 0)$. So for $|S| > -b$, the sum of the two losses is convex in $U_1$.
    
    Let $U_2\subset U_1$ be a compact neighborhood of $\vec p_y$ fully contained in the interior of $U_1$ such that the diameter of $U_2$ is smaller than the smallest distance of a point of $U_2$ to the boundary of $U_1$. Let $k_0$ be such that for $|S|>k_0$, the probability of the normalized histogram vector 
    \[\vec v_r = \pred S / |S|\] 
    being in $U_2$ is at least $p$. 
    
    Since $U_1$ is compact and $L_\mathrm{reg}<\infty$ on $U_1$, the regularization loss has a maximum value $m_{U_1}$. Let $k_1 > k_0$ be such that for $|S|>k_1$, the negative log-likelihood loss of a vector being more than the diameter of $U_1$ away from the minimum $v_r$ is more than $m_{U_1}$ higher than the negative log-likelihood loss at that minimum. 
    
    Conditioned on $\vec v_r \in U_2$, the combined loss at $v_r$ is smaller than at any point outside of $U_1$. As the combined loss is convex on $U_1$, there exists a unique global minimum. Since the event we conditioned in has probability at least $x$, so does the existence of a unique minimum.
    
    \item In the above, for any neighborhood $U_3\ni \vec p_y$, let $k_2>k_1$ be such that the probability of $v_r\in U_3\cap U_2$ is at least $x$. Then the global minimum is in $U_3$.
    
    \item The sum of the logarithms of likelihood and prior is minimized when the product of likelihood and prior is minimized, i.e. the posterior.
\end{enumerate}
\end{proof}

We propose to approximate this minimum and take it as estimator for the desired probabilities $\vec p_y$.

\begin{remark}\label{rem:global_urc}
~\begin{enumerate}
    \item We investigated different candidates for $L_\mathrm{reg}$. We settled on a loss proportional to the Kullback-Leibler divergence of the candidate $\vec p$ from an estimated for the distribution $\Prob[\dev]{Y}$, reasoning that the default assumption for the overall population should be one similar to the one observed in the training set.
    \item Although $|A|=n$ is necessary for solving the equation \refeq{eq:matrix_multplication} in lemma \ref{lemma:solvematrixeq}, it is not needed to optimize a function as in theorem \ref{thm:combinedloss}. It is perfectly reasonable to use partitions $A$ with more than $n$ elements\footnote{In our primary use case, we had good experiences with $n=2$, $|A|=4$.}.
\end{enumerate}
\end{remark}

\section{Local unsupervised recalibration}\label{sec:local_unsupervised_calibration} 
Assume that when applying your trained classifier in the field, you encounter different ``subpopulations'', each with their own probability distribution $P_{\app_1}, P_{\app_2}, \dots, P_{\app_s}$. In some cases it is reasonable to assume that the different subpopulations may be biased in different ways (for each $k$, the $\Prob[\app_k]{Y=i}$ values are different from $\Prob[\dev]{Y=i}$), although the relationship between $Y$ and $C$ is always the same in the field samples.

\begin{example}[Entomologist's classifier once again]
    Consider two ``subpopulations'' of insects: one of them have been caught close to the forest and the other close to the meadow.
    
    Maybe in forests beetles dominate while in meadow butterflies do, but the classifier cannot know this and account for this if not both groups were represented and recorded in the training set.
\end{example}

In that case, we can apply the global unsupervised recalibration procedure for each subpopulation individually, what describes Algorithm \ref{algo:urc_class}.

\begin{algorithm}[ht]
\label{algo:urc_class}
\SetAlgoLined
 partition predictions into sets using equation \refeq{eq:example_partition}\;
 \For{each category (if separate categories are encountered in the field)}{
    get posterior ground truth distribution by minimizing the loss given in definition \ref{def:Lll} plus an optional regularization loss\;
    \For{each sample in category}{
    obtain calibrated probability by applying lemma \ref{lemma:knownbias}\;
    }
 }
 \caption{Unsupervised recalibration for classification}
\end{algorithm}

For subpopulations with few encountered examples, we will stay mainly with our priors due to the regularization. For subpopulations with many examples, we will converge against the true values according to Theorem \ref{thm:combinedloss} -- the recalibrated classifier is then well calibrated for each subpopulation individually for which there is sufficient data.

\section{Extension to regression}\label{sec:regression}

As usual, a regression model that produces a probability distribution can be calibrated by splitting the support of the distribution in $n$ intervals. The regression model is equivalent to a probabilistic classifier that gives a probability for each interval and individual regression models that give distributions within each interval conditioned on the event that the true value is in that interval. The probabilistic classifier is then recalibrated as described above.

Since such a classifier predicts an ordered categorical set, it makes sense to include a continuity component into $L_\mathrm{reg}$, i.e. $L_\mathrm{reg}$ should generally increase if $|C_i -C_{i+1}| \gg 0$.

Also, it makes sense to choose a partition based on intervals of the predicted overall value (which is a linear combination of the $C_i$). Analogous to equation \refeq{eq:example_partition}, our suggestion would be to split into $n$ quantiles (as observed on the training data).

%% file: sec-experiment.tex
\section{Experiments}\label{sec:experiments}

Unsupervised calibration has the potential to improve a large range of probabilistic classifiers. To test this claim on a state-of-the-art classifier, we used Wolfram's ImageIdentify Net V1 \citep{ImageIdentify} to classify low resolution images.

The classification task was deciding whether a given image depicts a beetle or a butterfly\footnote{For the purpose of this article, we classified moths as butterflies as well, since together they comprise the order Lepidoptera. This avoids questions like ``Are skippers moths or butterflies?" and lets us compare one biological order (Lepidoptera) against another (Coleoptera). While there are many species of moths, the majority of our pictures of lepidopterans depict actual butterflies.}. These categories were chosen as typical, but visually distinct orders of insects for which there is a good supply of training data available. We obtained our data by decreasing the resolution of pictures from the iNaturalist Challenge dataset \citep{iNaturalist}. It comprises 57,742 pictures, each size reduced to six different sizes with a maximum dimension of 30, 40, 50, 75, 100 and 200 pixels respectively, while retaining the original aspect ratio.

The necessary code to obtain that dataset and replicate the following experiments is open-sourced in \citet{replicate-experiment}. 

Since the image classifier we use has been built as a general purpose image classifier not limited to beetles or butterflies, we remove all other predictions and re-normalise so that $p_\text{beetle}$ + $p_\text{butterfly} = 1$. We are aware that this is a crude way to force a prediction, but contend that this does not detract from this classifier's ability to serve as a proof-of-principle for the method under consideration.

The resulting classifier is not well calibrated even for a balanced training set, so we calibrate it first\footnote{We use Platt scaling \citep{platt-scaling}, which brings down the calibration component of the Brier score on a balanced set from $3\%$ to $0.1\%$.}. This benefits the classifier before recalibration more strongly than the classifier after recalibration. 

\begin{figure}[ht]
  \centering
  \includegraphics[width=\textwidth]{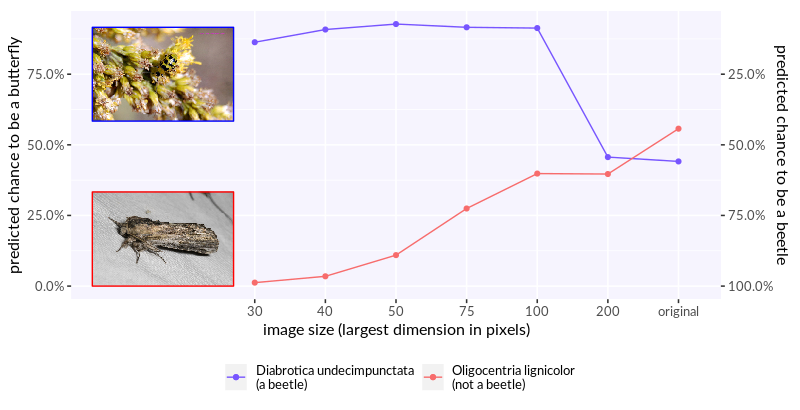}
  \caption{\small {\em Examples where the classifier needs high resolution to correctly solve the classification task.} The spotted cucumber beetle (above) only fills a small portion of the image and is crawling over buds which at low resolutions might conceivably resemble the folded wings of a butterfly. The white-streaked prominent (below) is a moth, which generally suffer from a higher misclassification rate. Its brown-grey color is more typical for beetles than for butterflies. The example pictures in the plot have been downscaled to contain 100 pixels in their largest dimension.}
  \label{fig:yyValidation}
 \end{figure}

At full resolution, beetles have a $86.1\%$ chance of being identified\footnote{
When evaluating the classifier as a hard classifier, we take as its prediction the class to which it assigns the higher probability.} 
correctly, and butterflies have a $87.7\%$ chance. It proved impossible to compare these numbers with the performance of ImageIdentify on its original training set, since neither those statistics nor the training set itself are published. However, such summaries (yielding $m_A$) are required as parameters for unsupervised calibration. To approximate them, we use a small balanced set of 200 randomly selected beetles and butterflies images at full resolution to simulate the evaluation of the classifier on the training set. We perform unsupervised calibration on the predictions of the classifier on the downsized versions of the other images.

\subsection{Global unsupervised recalibration}
For all tested image sizes, unsupervised calibration improved the log likelihood, Brier score and accuracy of the associated hard classifier considerably. This effect was strongest at low resolutions where the original classifier was weakest. It was robust\footnote{In  Table \ref{tab:global-experiment}, when running the set of 100 experiments with different numbers of partitions (2, 3, 8, and 16), results never differed from the reported values by more than relative $9\%$ for any value and more than relative $5\%$ for any value other than the 30 pixels one, except for the post-calibration calibration component of the Brier score, where the low absolute values make relative differences less relevant.} against the number of partitions and the choice of evaluation set. The results are shown in Table \ref{tab:global-experiment}. 

\begin{table}[ht]
    \centering
    \begin{tabular}{c||c|c|c|c|c|c}
            & \multicolumn{6}{c}{\makecell{image size in pixels}}\\
            & 30 & 40 & 50 & 75 & 100 & 200\\
           \hline
           \hline
         \makecell{negative log likelihood\\per sample} &
            \makecell{0.399 \\to\\ 0.357} &
            \makecell{0.378 \\to\\ 0.300} &
            \makecell{0.359 \\to\\ 0.276} &
            \makecell{0.323 \\to\\ 0.242} &
            \makecell{0.307 \\to\\ 0.220} &
            \makecell{0.286 \\to\\ 0.184} \\
         \hline
         Brier score & 
            \makecell{0.126 \\to\\ 0.091} &
            \makecell{0.118 \\to\\ 0.083} &
            \makecell{0.111 \\to\\ 0.078} &
            \makecell{0.099 \\to\\ 0.070} &
            \makecell{0.095 \\to\\ 0.064} &
            \makecell{0.087 \\to\\ 0.055} \\
         \hline
         \makecell{calibration component\\ of Brier score} &
            \makecell{0.044 \\to\\ 0.008} &
            \makecell{0.041 \\to\\ 0.004} &
            \makecell{0.037 \\to\\ 0.004} &
            \makecell{0.032 \\to\\ 0.003} &
            \makecell{0.032 \\to\\ 0.002} &
            \makecell{0.032 \\to\\ 0.001} \\
         \hline
         hard classification accuracy & 
            \makecell{0.826 \\to\\ 0.893} &
            \makecell{0.837 \\to\\ 0.898} &
            \makecell{0.847 \\to\\ 0.902} &
            \makecell{0.863 \\to\\ 0.910} &
            \makecell{0.869 \\to\\ 0.916} &
            \makecell{0.879 \\to\\ 0.926} \\
    \end{tabular}
    
      \caption{\small {\em Effects of global unsupervised recalibration}. All numbers have been averaged over 100 different random choices of the evaluation set and were calculated using 4 partitions.}
    \label{tab:global-experiment}
\end{table}

\begin{table}
\end{table}

\begin{figure}[ht]
  \centering
  \includegraphics[width=\textwidth]{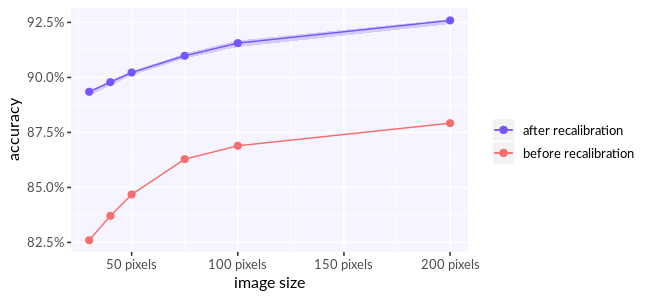}
  \caption{\small {\em Hard classification accuracy for the original and recalibrated classifier.} The ribbons around the blue dots represent the $95\%$ range for the values depending on different choices of evaluation set. Recalibrating the classifier on the low 30 pixel resolution still produces a more accurate result than not recalibrating the classifier on the relatively high resolution of 200 pixels.}
  \label{fig:global-experiment-accuracy}
\end{figure}

In all cases, unsupervised recalibration computes that the base rate for beetles is at most $17\%$ (it is actually $11\%$), while the average unrecalibrated probability was between $23\%$ and $27\%$. Accordingly, recalibration reclassifies some images that were previously considered beetles as butterflies. This increases the average precision for the beetles predictions considerably ($32\%$ to $63\%$ for 30 pixel images, and $47\%$ to $76\%$ for 200 pixel images). Conversely, the average precision for butterfly predictions \textit{decreases} only slightly ($94\%$ to $90\%$ for 30 pixel images, and $98\%$ to $94\%$ for 200 pixel images). In all 100 experiments, the classification accuracy increases by at least $6\%$ (mean increase: $7\%$) for 30 pixel images, and by at least $4\%$ (mean increase: $5\%$) for 200 pixel images (see Figure \ref{fig:global-experiment-accuracy}).

A good way to evaluate the performance of a probabilistic classifier is the Brier score \citep{brier}. This score can be decomposed\footnote{The decomposition requires a choice for partition of the predictions. The numbers we report here have been computed using deciles.} into a refinement and a calibration component. Intuitively speaking, refinement measures the classifier's ability to distinguish between samples which are highly likely to belong to one class and samples which are highly likely to belong to the other class, while the calibration component measures that these likelihoods are reported correctly.

\begin{figure}[ht]
  \centering
  \includegraphics[width=\textwidth]{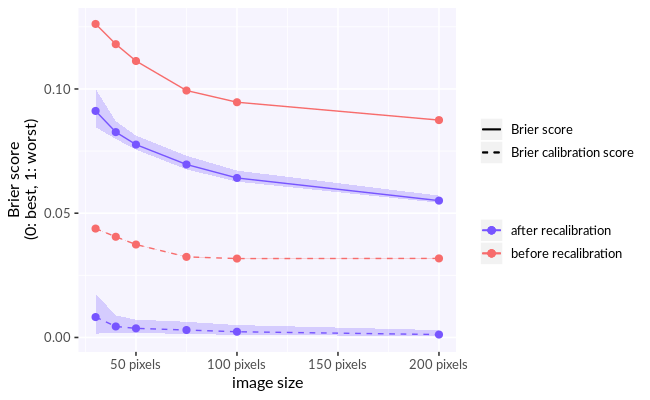}
  \caption{\small {\em Brier score before and after recalibration.} The Brier score is composed of the refinement component, which is unaffected by global unsupervised recalibration, and the calibration component, which decreases considerably under recalibration. The ribbons around the blue dots represent the $95\%$ range for the values depending on different choices of evaluation set.}
  \label{fig:global-experiment-Brier}
\end{figure}

Figure \ref{fig:global-experiment-Brier} shows the improvement in the Brier score. Global unsupervised recalibration does not impact a classifier's refinement, so all improvement in the Brier score is due to the improvement in its calibration component. This is in stark contrast to local unsupervised recalibration (see below).

\subsection{Local unsupervised recalibration}

Local unsupervised recalibration takes advantage of the division into subpopulations. In this case, this could entail sorting the images by identity of the photographer, or the location or season in which they were created. It is likely that restricted to each class, the unmodified classifier works similarly well on all these subpopulations, yet the different subpopulations probably have different base rates for the classes. This difference in base rates determines the strength of local unsupervised recalibration.

We want to test this relationship systematically from a neutral starting point and separate it from the global effect. So we use the full resolution images and subset the iNaturalist training data to make it balanced. Since we made sure to start with a classifier that is already well calibrated in this setting, there is nothing for global unsupervised recalibration to improve here.

\begin{figure}[t!]
  \centering
  \includegraphics[width=\textwidth]{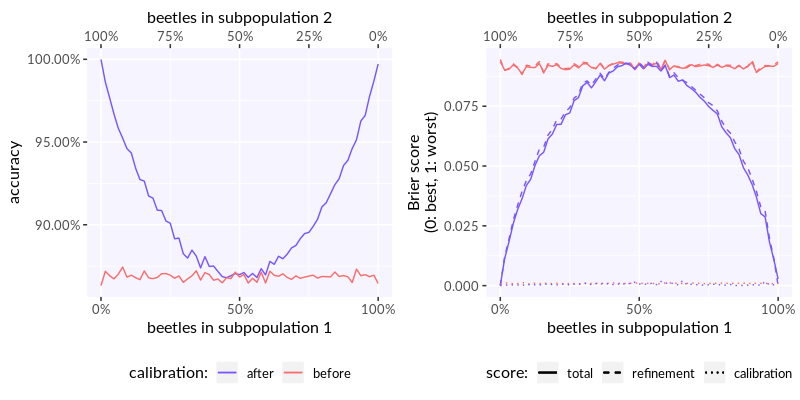}
  \caption{\small {\em Effect of local unsupervised recalibration.} Local unsupervised recalibration is most beneficial if the subpopulation differ substantially in their class membership distributions. Recalibration shown for 4 partitions, but other numbers yield highly similar results. Since the classifier was well calibrated from the beginning, the calibration component of the Brier score is close to 0, and the refinement component is close to the total Brier score.}
  \label{fig:local-experiment}
\end{figure}

We then randomly assign the remaining 12,894 images to subpopulations $1$ and $2$ such that the number of beetles contained in each corresponds to a set base rate. It turns out that unsupervised recalibration never hurts\footnote{In other examples, however, it can hurt if the classifier started out being biased for the subpopulations, or if the amount of data is so small that the estimated base rate is unluckily far from the actual base rate.}, with its benefits being strongest for very unbalanced subpopulations (see Figure \ref{fig:local-experiment}).

In contrast to global recalibration, local recalibration improves the \textit{refinement} of the classifier. Since we started with an already well calibrated classifier, the calibration component of the Brier score is always (close to) 0, while the refinement component only approaches 0 if subpopulation membership completely determines class membership.

This strongly underlines than when natural subpopulations occur which are not expected to be independent from the quantity one wishes to predict, local unsupervised recalibration is highly effective. If it is impossible or unfeasible to retrain the classifier with the subpopulation as an input feature, unsupervised recalibration has the potential to improve performance \textit{considerably}.

\subsection{Quantification}
The dataset shift described above relies on the \emph{quantification} task, that is the detection of the true prevalence in the test data set (cf. Lemma \ref{lemma:knownbias}). We compare Unsupervised Recalibration (URC) with three standard algorithms: 
\begin{enumerate}
    \item Classify and Count (CC), described e.g. in \citet[\textsection 2.1]{karpov-etal-2016-nru}
    \item Adjusted Classify and Count (ACC), introduced by \citet{gart-buck} and independently rediscovered by \cite{Saerens-2001-adjustingtheoutputs}, and \cite{forman},
    \item Expectation Maximization (EM), introduced by \cite{peters-coberly} and rediscovered by \citet{Saerens-2001-adjustingtheoutputs}.
\end{enumerate}

For a comprehensive review of quantification algorithms see \citet[\textsection 2]{karpov-etal-2016-nru}, \citet[\textsection\textsection 6--8]{gonzalez-review-quantification} or \citet[\textsection 3]{tasche-fisher-consistency}. Similarly to Unsupervised Recalibration, these algorithms do not require the classifier to be retrained multiple times. For this reason, we do not consider the popular CDE-Iterate algorithm \citep{xue-weiss}.

\paragraph{Experimental setup} We test the quantification algorithms on artificial data sets with binary labels constructed using the manner described in \citet[\textsection 3.1]{karpov-etal-2016-nru}. Since this is a random procedure, we run 30 replicas for each data point. The whole experiment is described as Algorithm \ref{algo:comparison_exp}.

\begin{algorithm}[tb]
\label{algo:comparison_exp}
\algorithmfootnote{
${}^*$ Both Unsupervised Recalibration and Adjusted Classify and Count rely on the knowledge about the classifier confidence matrix. We estimate it on a sufficiently large validation data set, which prevalence is the same as in the training data set. In reality, big validation data sets may not be available, and this can be replaced with many-fold cross-validation.\\
${}^{**}$ Adjusted Classify and Count has the access to the confidence matrix calculated in the last step.\\
${}^{***}$ Unsupervised Recalibration uses an analogue of the confidence matrix, but for a given partition, as described in \textsection \ref{sec:global_unsupervised_calibration}.\\
${}^{****}$ Expectation Maximization uses the true prevalence on the training data set as a starting point.
}
\SetAlgoLined
    \Switch{Experiment}{
        \uCase{balanced training experiment}{
            prevalence in test = 5\%, prevalence in training = 50\%
        }
        \uCase{balanced test experiment}{
            prevalence in test = 50\%, prevalence in training = 5\%
        }}
 size of training data = 2000\;
 \For{size of test data = 50, 100, 500, 1000, 3000}{
    \For{replica = 1 ... 30}{
        generate binary data set $D$\;
        split into training $D^{train}$, validation$^*$ $D^{valid}$, and test data $D^{test}_i$ sets of given prevalence and size\;
        train a logistic regression classifier $Cl_i$ on $D_i^{train}$, and calculate its predictions on $D_i^{valid}$\;
        apply the classifier $Cl_i$ to $D_i^{test}$ and apply each of four quantification algorithms$^{**, ***, ****}$ to get estimated prevalence
    }
    aggregate estimated prevalences\;
    \nl\KwRet{distribution for estimated prevalence}
    }
 \caption{Comparison with state of the art dataset shift algorithms.}
\end{algorithm}

This experiment produces for each data set size and quantification algorithm an empirical distribution for the estimated prevalence. More detailed information can be found in Appendix \ref{section:appendix-quantification}. Code which can be used to reproduce our results is available in the repository \citet{replicate-experiment}.

\begin{figure}[t!]
  \centering
  \includegraphics[width=\textwidth]{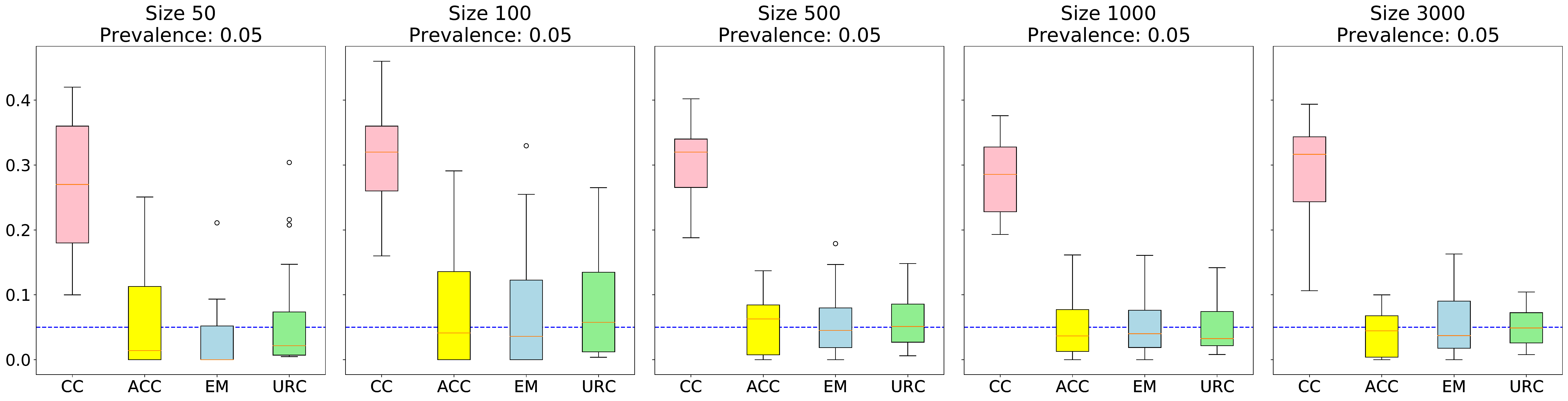}
  \caption{\small {\em Balanced training data set experiment}. The training data set is balanced and has fixed size. The size of the test data set varies between $50$ and $3,000$ samples. Blue dotted line represents the true $5\%$ prevalence of the test data set.}
  \label{fig:quantification-training-row}
\end{figure}

\paragraph{Balanced training data set}
In Figure \ref{fig:quantification-training-row} we present the results of five experiments. In each of them, the training (and validation) data set are balanced and have fixed size of 2,000 samples. The test data set has prevalence of $5\%$ and its size varies between $50$ and $3,000$ samples.

This experiment does not show a significant performance difference between Adjusted Classify and Count, Expectation Maximization and Unsupervised Recalibration. The naive version of Classify and Count does not recover the true prevalence.

\begin{figure}[t!]
  \centering
  \includegraphics[width=\textwidth]{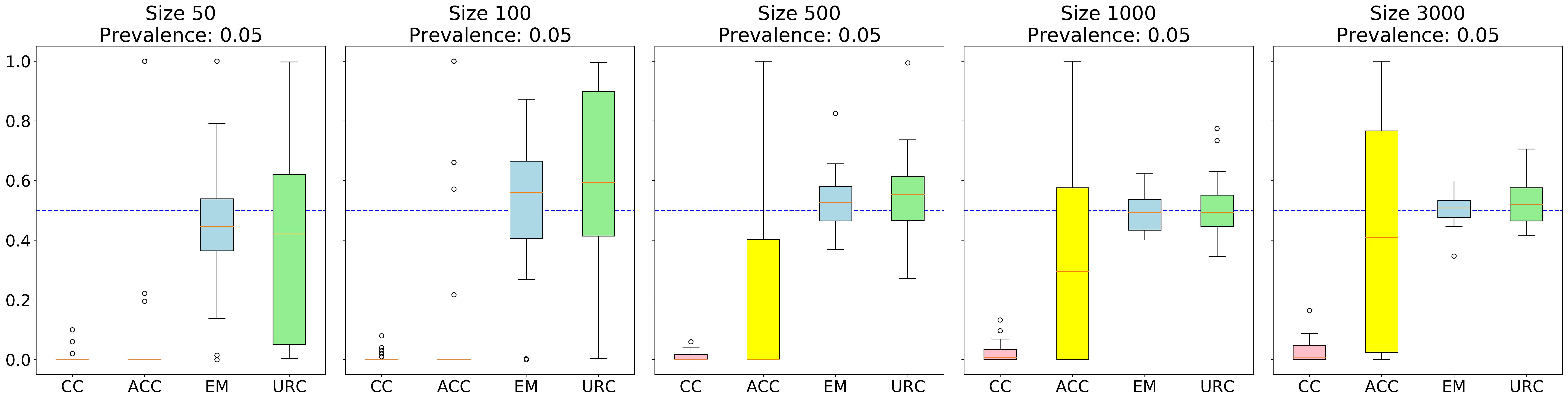}
    \caption{\small {\em Balanced test data set experiment}. The training data set has fixed size but has the prevalence of $5\%$. The test data set is balanced and its size varies between $50$ and $3,000$ samples. Blue dotted line represents the true $50\%$ prevalence of the test data set.}
  \label{fig:quantification-test-row}
\end{figure}

\paragraph{Balanced test data set}
In Figure \ref{fig:quantification-test-row} we present the results of five experiments. In each of them, the training (and validation) data set have fixed size of 2,000 samples and the prevalence of $5\%$. The test data set is balanced and its size varies between $50$ and $3,000$ samples.

The performance of Unsupervised Recalibration and Expectation Maximization converges to the true value, while the Adjusted and naive versions of Classify and Count do not reliably recover the true prevalence.

Presented figures are truncated versions of Figures \ref{fig:quantification-training-grid} and \ref{fig:quantification-test-grid} from Appendix \ref{section:appendix-quantification}. The experimental results suggest that Unsupervised Recalibration may be treated as a plausible alternative to Expectation Maximization. Expectation Maximization should however be preferred if the validation data set (or computing power for cross-validation) is not available. On the other hand, the confidence matrix available to Unsupervised Recalibration makes it possible to go on to bound the uncertainty associated with the detected prevalence.

%% file: sec-afterword.tex
\section{Contraindications}
\label{sec:counterindications}

The technique described in this article cannot be applied blindly. In particular, there are two big contraindications that should always be considered carefully. Do not apply local recalibration if any of the following holds:

\begin{enumerate}
    \item The original classifier has a bias for the subpopulations under consideration -- this bias would be increased with local unsupervised recalibration.

    Such a bias commonly arises from the original classifier having access to features which are a good proxy for the subpopulation. It could also arise from the training data labels being tainted \citep{label-bias}.
    
    \item The classification is desired to be bias free for the subpopulations under consideration -- local unsupervised recalibration will introduce such a bias.
    
    The original classifier might not take the subpopulation into account by design. E.g. while parental income might be correlated with academic success, it is strongly contraindicated to recalibrate university admissions tests by parental income bracket, which would have the effect of preferring the more affluent applicant in the case of similar objective scores. 
\end{enumerate}

Additionally, it is sometimes considered desirable in a hard classifier to have similar performance on all classes. Unsupervised recalibration does not optimize for this property -- in fact, it deliberately sacrifices performance on rare classes to gain improved performance on common classes. 

However, if the consistency assumption (Axiom \ref{ax:consistency}) appears plausible, it is still advisable to run unsupervised recalibration in order to determine the base rate $P(Y = i)$, which allows to transform the probabilistic classifier into a hard classifier in a way such that performance on all classes is maximised.

\section{Summary}
Unsupervised recalibration addresses two common problems in applying machine learning models:

\begin{itemize}
    \item A model is applied in an environment where the ground truth distribution is not guaranteed to reflect the distribution in the model's training set (i.e., the training set may exhibit an unknown bias).
    \item During model application, samples can be sorted into relevant subpopulations which were not taken into account to train the model (i.e., new features become available).
\end{itemize}

In these situations, unsupervised recalibration can improve classification results by a considerable margin (see sections \ref{sec:global_unsupervised_calibration} and \ref{sec:local_unsupervised_calibration}). In contrast to established methods, it does not require gathering new ground truth for the new environment or subpopulations, which is often extremely costly or impossible, and without retraining the original ML model, which is sometimes costly and often impossible.

\section*{Acknowledgement}
We would like to thank Ian Wright for valuable comments on the manuscript and GitHub Inc. for supporting the research.

%% file: sec-appendix.tex
\section{Quantification experiment details}
\label{section:appendix-quantification}

We implemented the experiments in Python \citep{python} using \texttt{scikit-learn} \citep{scikit-learn}, \texttt{SciPy} ecosystem \citep{scipy}, and \texttt{pytorch} \citep{pytorch}. For Unsupervised Recalibration we used a partition into two intervals, split by the median of the predictions on the validation data set.

Every training and validation data set we generated consisted of 2,000 data samples. We generated the data set using the \texttt{make\_classification} function with $\texttt{class\_sep}=0.4$ and $\texttt{flip\_y}=0.1$. Each data point had four features two of which were irrelevant for the problem. 

Expectation Maximization and Unsupervised Recalibration are iterative algorithms. To ensure that Expectation Maximization and Unsupervised Recalibration have converged, we ran each experiment increasing the number of optimization steps from $3,000$ to $6,000$ for Expectation Maximization, and from $5,000$ to $10,000$ for Unsupervised Recalibration. There was no visible difference between the generated figures.

\begin{figure}[t!]
  \centering
  \includegraphics[width=0.9\textwidth]{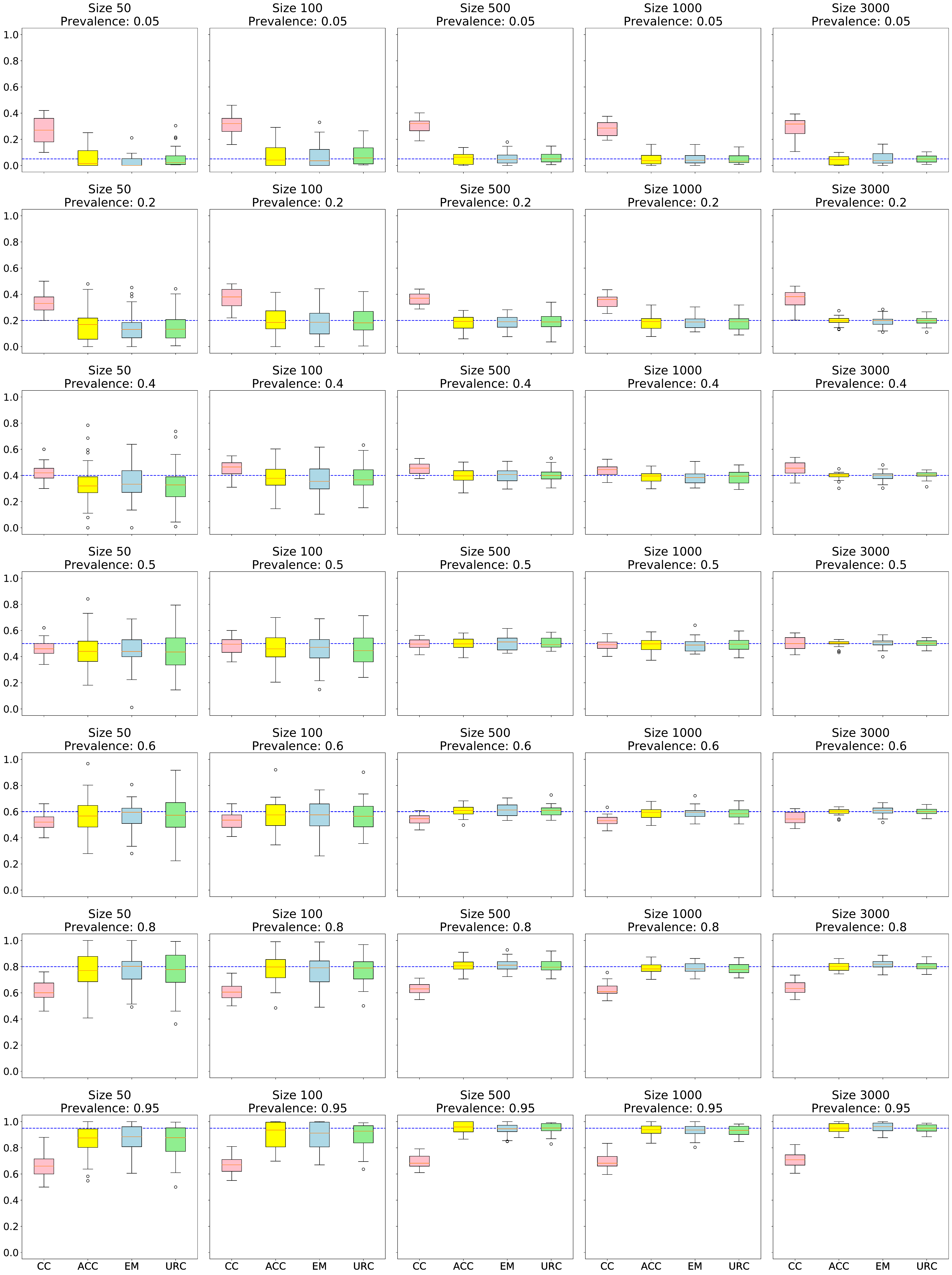}
  \caption{\small {\em In this experiment the training data set is balanced and we do not change its size. We change the test data set size and its prevalence (marked with a horizontal dotted line). Apart from naive Classify and Count, performance of all algorithms is similar and improves when the test data set size is increased, converging to the true value.}}
  \label{fig:quantification-training-grid}
\end{figure}

\begin{figure}[t!]
  \centering
  \includegraphics[width=0.9\textwidth]{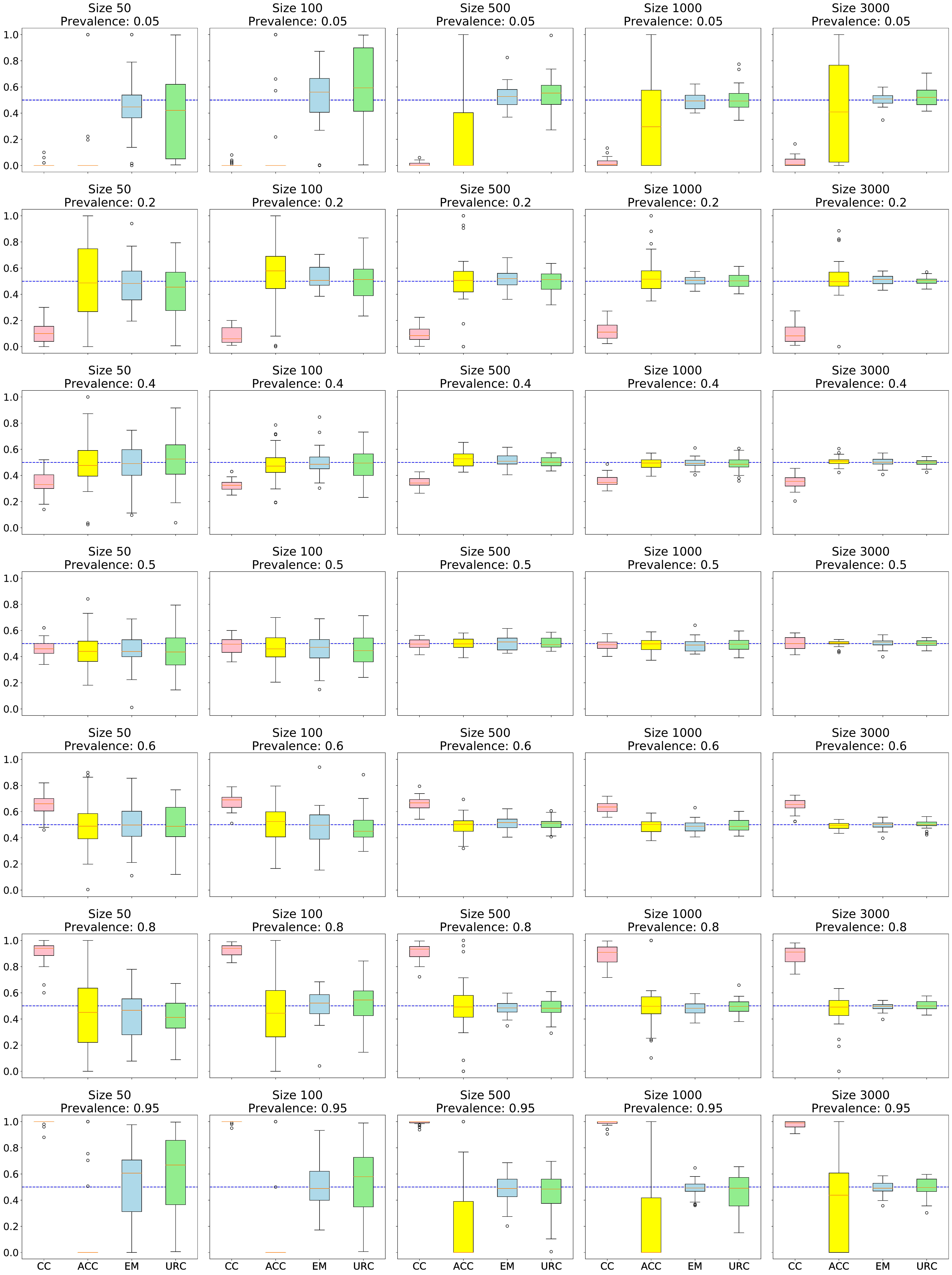}
  \caption{\small {\em In this experiment we change the size of the test data set keeping it balanced, what is symbolized by the horizontal dotted line. The training data set has varying prevalence. Classify and Count, whether adjusted or not, is not reliable when there is a big mismatch between prevalence of the training and test data sets. Expectation Maximization and Unsupervised Recalibration converge to the true prevalence for sufficiently large test data sets.}}
  \label{fig:quantification-test-grid}
\end{figure}

As mentioned in \textsection \ref{sec:experiments}, Figures \ref{fig:quantification-training-row} and \ref{fig:quantification-test-row} should be treated as truncated versions of Figures \ref{fig:quantification-training-grid} and \ref{fig:quantification-test-grid}, respectively.

%% file: main.bbl
\begin{thebibliography}{29}
\providecommand{\natexlab}[1]{#1}
\providecommand{\url}[1]{\texttt{#1}}
\expandafter\ifx\csname urlstyle\endcsname\relax
  \providecommand{\doi}[1]{doi: #1}\else
  \providecommand{\doi}{doi: \begingroup \urlstyle{rm}\Url}\fi

\bibitem[{Arjovsky} et~al.(2019){Arjovsky}, {Bottou}, {Gulrajani}, and
  {Lopez-Paz}]{invariant-risk-minimization}
Martin {Arjovsky}, L{\'e}on {Bottou}, Ishaan {Gulrajani}, and David
  {Lopez-Paz}.
\newblock {Invariant Risk Minimization}.
\newblock \emph{arXiv e-prints}, art. arXiv:1907.02893, Jul 2019.

\bibitem[Bassett and Deride(2018)]{maximumaposteriori}
Robert Bassett and Julio Deride.
\newblock Maximum a posteriori estimators as a limit of bayes estimators.
\newblock \emph{Mathematical Programming}, pages 1--16, 2018.

\bibitem[Forman(2008)]{forman}
George Forman.
\newblock Quantifying counts and costs via classification.
\newblock \emph{Data Mining and Knowledge Discovery}, 17:\penalty0 164--206,
  October 2008.

\bibitem[Gart and Buck(1966)]{gart-buck}
J.J Gart and A.A Buck.
\newblock Comparison of a screening test and a reference test in epidemiologic
  studies. ii. a probabilistic model for the comparison of diagnostic tests.
\newblock \emph{American Journal of Epidemiology}, 83:\penalty0 593--602, May
  1966.

\bibitem[Gebel(2009)]{calibration}
Martin Gebel.
\newblock \emph{Multivariate calibration of classifier scores into the
  probability space}.
\newblock PhD thesis, University of Dortmund, 2009.

\bibitem[González et~al.(2017)González, Castaño, Chawla, and del
  Coz]{gonzalez-review-quantification}
Pablo González, Alberto Castaño, Nitesh Chawla, and Juan del Coz.
\newblock A review on quantification learning.
\newblock \emph{ACM Computing Surveys}, 50:\penalty0 1--40, 09 2017.
\newblock \doi{10.1145/3117807}.

\bibitem[Hern{\'a}ndez-Orallo et~al.(2011)Hern{\'a}ndez-Orallo, Flach, and
  Ramirez]{brier}
Jos{\'e} Hern{\'a}ndez-Orallo, Peter~A Flach, and C{\`e}sar~Ferri Ramirez.
\newblock Brier curves: a new cost-based visualisation of classifier
  performance.
\newblock In \emph{ICML}, pages 585--592, 2011.

\bibitem[Jiang and Nachum(2019)]{label-bias}
Heinrich Jiang and Ofir Nachum.
\newblock Identifying and correcting label bias in machine learning.
\newblock \emph{arXiv preprint arXiv:1901.04966}, 2019.

\bibitem[Karpov et~al.(2016)Karpov, Porshnev, and
  Rudakov]{karpov-etal-2016-nru}
Nikolay Karpov, Alexander Porshnev, and Kirill Rudakov.
\newblock {NRU}-{HSE} at {S}em{E}val-2016 task 4: Comparative analysis of two
  iterative methods using quantification library.
\newblock In \emph{Proceedings of the 10th International Workshop on Semantic
  Evaluation ({S}em{E}val-2016)}, pages 171--177, San Diego, California, June
  2016. Association for Computational Linguistics.
\newblock \doi{10.18653/v1/S16-1025}.
\newblock URL \url{https://www.aclweb.org/anthology/S16-1025}.

\bibitem[Kull et~al.(2017)Kull, Silva~Filho, and Flach]{beta-calibration}
Meelis Kull, Telmo Silva~Filho, and Peter Flach.
\newblock Beta calibration: a well-founded and easily implemented improvement
  on logistic calibration for binary classifiers.
\newblock In \emph{Artificial Intelligence and Statistics}, pages 623--631,
  2017.

\bibitem[Lipton et~al.(2018)Lipton, Wang, and Smola]{lipton2018detecting}
Zachary~C Lipton, Yu-Xiang Wang, and Alex Smola.
\newblock Detecting and correcting for label shift with black box predictors.
\newblock \emph{arXiv preprint arXiv:1802.03916}, 2018.

\bibitem[Moreno-Torres et~al.(2012)Moreno-Torres, Raeder, Alaiz-Rodríguez,
  Chawla, and Herrera]{Moreno-Torres}
Jose~G. Moreno-Torres, Troy Raeder, Rocío Alaiz-Rodríguez, Nitesh~V. Chawla,
  and Francisco Herrera.
\newblock A unifying view on dataset shift in classification.
\newblock \emph{Pattern Recognition}, 45\penalty0 (1):\penalty0 521 -- 530,
  2012.
\newblock ISSN 0031-3203.
\newblock \doi{https://doi.org/10.1016/j.patcog.2011.06.019}.
\newblock URL
  \url{http://www.sciencedirect.com/science/article/pii/S0031320311002901}.

\bibitem[Paszke et~al.(2019)Paszke, Gross, Massa, Lerer, Bradbury, Chanan,
  Killeen, Lin, Gimelshein, Antiga, Desmaison, Kopf, Yang, DeVito, Raison,
  Tejani, Chilamkurthy, Steiner, Fang, Bai, and Chintala]{pytorch}
Adam Paszke, Sam Gross, Francisco Massa, Adam Lerer, James Bradbury, Gregory
  Chanan, Trevor Killeen, Zeming Lin, Natalia Gimelshein, Luca Antiga, Alban
  Desmaison, Andreas Kopf, Edward Yang, Zachary DeVito, Martin Raison, Alykhan
  Tejani, Sasank Chilamkurthy, Benoit Steiner, Lu~Fang, Junjie Bai, and Soumith
  Chintala.
\newblock Pytorch: An imperative style, high-performance deep learning library.
\newblock In H.~Wallach, H.~Larochelle, A.~Beygelzimer, F.~d'~Alch\'{e}-Buc,
  E.~Fox, and R.~Garnett, editors, \emph{Advances in Neural Information
  Processing Systems 32}, pages 8024--8035. Curran Associates, Inc., 2019.
\newblock URL
  \url{http://papers.neurips.cc/paper/9015-pytorch-an-imperative-style-high-performance-deep-learning-library.pdf}.

\bibitem[Pedregosa et~al.(2011)Pedregosa, Varoquaux, Gramfort, Michel, Thirion,
  Grisel, Blondel, Prettenhofer, Weiss, Dubourg, Vanderplas, Passos,
  Cournapeau, Brucher, Perrot, and Duchesnay]{scikit-learn}
F.~Pedregosa, G.~Varoquaux, A.~Gramfort, V.~Michel, B.~Thirion, O.~Grisel,
  M.~Blondel, P.~Prettenhofer, R.~Weiss, V.~Dubourg, J.~Vanderplas, A.~Passos,
  D.~Cournapeau, M.~Brucher, M.~Perrot, and E.~Duchesnay.
\newblock Scikit-learn: Machine learning in {P}ython.
\newblock \emph{Journal of Machine Learning Research}, 12:\penalty0 2825--2830,
  2011.

\bibitem[Peters and Coberly(1976)]{peters-coberly}
C.~Peters and W.A Coberly.
\newblock The numerical evaluation of the maximum-likelihood estimate of
  mixture proportions.
\newblock \emph{Communications in Statistics -- Theory and Methods},
  5:\penalty0 1127--1135, 1976.

\bibitem[Platt et~al.(1999)]{platt-scaling}
John Platt et~al.
\newblock Probabilistic outputs for support vector machines and comparisons to
  regularized likelihood methods.
\newblock \emph{Advances in large margin classifiers}, 10\penalty0
  (3):\penalty0 61--74, 1999.

\bibitem[Saerens et~al.(2001)Saerens, Latinne, and
  Decaestecker]{Saerens-2001-adjustingtheoutputs}
Marco Saerens, Patrice Latinne, and Christine Decaestecker.
\newblock Adjusting the outputs of a classifier to new a priori probabilities:
  A simple procedure.
\newblock \emph{Neural Computation}, 14:\penalty0 14--21, 2001.

\bibitem[S{\"a}rndal et~al.(2003)S{\"a}rndal, Swensson, and
  Wretman]{stratification}
Carl-Erik S{\"a}rndal, Bengt Swensson, and Jan Wretman.
\newblock \emph{Model assisted survey sampling}.
\newblock Springer Science \& Business Media, 2003.

\bibitem[Shahrokh~Esfahani and Dougherty(2013)]{biasthroughstratification}
Mohammad Shahrokh~Esfahani and Edward~R Dougherty.
\newblock Effect of separate sampling on classification accuracy.
\newblock \emph{Bioinformatics}, 30\penalty0 (2):\penalty0 242--250, 2013.

\bibitem[{Tasche}(2017)]{tasche-fisher-consistency}
Dirk {Tasche}.
\newblock {Fisher consistency for prior probability shift}.
\newblock \emph{arXiv e-prints}, art. arXiv:1701.05512, January 2017.

\bibitem[Van~Horn et~al.(2018)Van~Horn, Mac~Aodha, Song, Cui, Sun, Shepard,
  Adam, Perona, and Belongie]{iNaturalist}
Grant Van~Horn, Oisin Mac~Aodha, Yang Song, Yin Cui, Chen Sun, Alex Shepard,
  Hartwig Adam, Pietro Perona, and Serge Belongie.
\newblock The inaturalist species classification and detection dataset.
\newblock In \emph{The IEEE Conference on Computer Vision and Pattern
  Recognition (CVPR)}, June 2018.

\bibitem[Van~Rossum and Drake~Jr(1995)]{python}
Guido Van~Rossum and Fred~L Drake~Jr.
\newblock \emph{Python reference manual}.
\newblock Centrum voor Wiskunde en Informatica Amsterdam, 1995.

\bibitem[Vaz et~al.(2019)Vaz, Izbicki, and Stern]{Vaz-Izbicki-Stern}
Afonso~Fernandes Vaz, Rafael Izbicki, and Rafael~Bassi Stern.
\newblock Quantification under prior probability shift: the ratio estimator and
  its extensions.
\newblock \emph{Journal of Machine Learning Research}, 20\penalty0
  (79):\penalty0 1--33, 2019.
\newblock URL \url{http://jmlr.org/papers/v20/18-456.html}.

\bibitem[{Virtanen} et~al.(2020){Virtanen}, {Gommers}, {Oliphant}, {Haberland},
  {Reddy}, {Cournapeau}, {Burovski}, {Peterson}, {Weckesser}, {Bright}, {van
  der Walt}, {Brett}, {Wilson}, {Jarrod Millman}, {Mayorov}, {Nelson}, {Jones},
  {Kern}, {Larson}, {Carey}, {Polat}, {Feng}, {Moore}, {Vand erPlas},
  {Laxalde}, {Perktold}, {Cimrman}, {Henriksen}, {Quintero}, {Harris},
  {Archibald}, {Ribeiro}, {Pedregosa}, {van Mulbregt}, and
  {Contributors}]{scipy}
Pauli {Virtanen}, Ralf {Gommers}, Travis~E. {Oliphant}, Matt {Haberland}, Tyler
  {Reddy}, David {Cournapeau}, Evgeni {Burovski}, Pearu {Peterson}, Warren
  {Weckesser}, Jonathan {Bright}, St{\'e}fan~J. {van der Walt}, Matthew
  {Brett}, Joshua {Wilson}, K.~{Jarrod Millman}, Nikolay {Mayorov}, Andrew
  R.~J. {Nelson}, Eric {Jones}, Robert {Kern}, Eric {Larson}, CJ~{Carey},
  {\.I}lhan {Polat}, Yu~{Feng}, Eric~W. {Moore}, Jake {Vand erPlas}, Denis
  {Laxalde}, Josef {Perktold}, Robert {Cimrman}, Ian {Henriksen}, E.~A.
  {Quintero}, Charles~R {Harris}, Anne~M. {Archibald}, Ant{\^o}nio~H.
  {Ribeiro}, Fabian {Pedregosa}, Paul {van Mulbregt}, and SciPy 1.~0
  {Contributors}.
\newblock {SciPy 1.0: Fundamental Algorithms for Scientific Computing in
  Python}.
\newblock \emph{Nature Methods}, 17:\penalty0 261--272, 2020.
\newblock \doi{https://doi.org/10.1038/s41592-019-0686-2}.

\bibitem[Wolfram(2017)]{ImageIdentify}
Stephen Wolfram.
\newblock Wolfram {I}mage{I}dentify {N}et {V}1.
\newblock
  \url{https://resources.wolframcloud.com/NeuralNetRepository/resources/Wolfram-ImageIdentify-Net-V1},
  2017.

\bibitem[Xue and Weiss(2009)]{xue-weiss}
Jack Xue and Gary Weiss.
\newblock Quantification and semi-supervised classification methods for
  handling changes in class distribution.
\newblock pages 897--906, 01 2009.
\newblock \doi{10.1145/1557019.1557117}.

\bibitem[Ying(2019)]{overfitting}
Xue Ying.
\newblock An overview of overfitting and its solutions.
\newblock \emph{Journal of Physics: Conference Series}, 1168:\penalty0 022022,
  feb 2019.
\newblock \doi{10.1088/1742-6596/1168/2/022022}.
\newblock URL \url{https://doi.org/10.1088%2F1742-6596%2F1168%2F2%2F022022}.

\bibitem[Ziegler and Czy\.z(2019)]{replicate-experiment}
Albert Ziegler and Pawe\l\ Czy\.z.
\newblock Unsupervised recalibration experiments.
\newblock \url{https://github.com/albert-ziegler/unsupervised-calibration},
  2019.

\bibitem[Zitkovic(2013)]{conditional-expectation}
Gordan Zitkovic.
\newblock Conditional expectation.
\newblock
  \url{https://web.ma.utexas.edu/users/gordanz/notes/conditional_expectation.pdf},
  2013.

\end{thebibliography}
